\documentclass[]{amsart} 

\usepackage{graphicx}
\usepackage[T1]{fontenc} 
\usepackage{ae, aecompl} 
\usepackage{booktabs}
\usepackage{fix-cm}

\usepackage{amssymb}
\usepackage{amsmath}
\usepackage{longtable,verbatim}
\usepackage{booktabs}
\usepackage{color, colortbl}
\usepackage{algorithm}
\usepackage{algorithmic}
\usepackage{subfigure}
\usepackage{url}
\usepackage{multirow} 
\begin{document}

\title[DBMs in EDAs for Combinatorial Optimization]{Deep Boltzmann Machines in Estimation of Distribution Algorithms for Combinatorial Optimization}

\author[]{Malte Probst}
\author[]{Franz Rothlauf}
\address[]{Johannes Gutenberg-Universität Mainz\\
Dept.~of Information Systems and Business Administration\\
Jakob-Welder-Weg 9, 55128 Mainz, Germany}
\email{\{probst|rothlauf\}@uni-mainz.de}
\urladdr{http://wi.bwl.uni-mainz.de}
\begin{abstract}
Estimation of Distribution Algorithms (EDAs) require flexible probability models that can be efficiently learned and sampled. Deep Boltzmann Machines (DBMs) are generative neural networks with these desired properties. We integrate a DBM into an EDA and evaluate the performance of this system in solving combinatorial optimization problems with a single objective. We compare the results to the Bayesian Optimization Algorithm. The performance of DBM-EDA was superior to BOA for difficult additively decomposable functions which are separable, i.e., concatenated deceptive traps of higher order. For most other benchmark problems, DBM-EDA cannot clearly outperform BOA, or other neural network-based EDAs. In particular, it often yields optimal solutions for a subset of the runs (with fewer evaluations than BOA), but is unable to provide reliable convergence to the global optimum competitively. At the same time, the model building process is fast, but computationally more expensive than that of other EDAs using probabilistic models from the neural network family, such as DAE-EDA. 
\end{abstract}


\maketitle

\section{Introduction}
\label{intro}
Estimation of Distribution Algorithms  (EDAs) \cite{Muehlenbein1996,larranaga2002estimation} are metaheuristics for combinatorial and continuous non-linear optimization. The maintain a population of candidate solutions to the optimization problem at hand (see Algorithm \ref{alg-eda}). First, they select solutions with a high quality from the population. Subsequently, they build a model that approximates the probability distribution of these solutions. Then, new candidate solutions are sampled from the model. The EDA then starts over by selecting the next set of good solutions from the new candidate solutions and the previous selection.

In order to be suitable for an EDA, a model therefore has to fulfill certain criteria:
\begin{itemize}
 \item It must be able to approximate the probability distribution of the selected individuals. 
 \item It must be able to sample new solutions from this probability distribution, serving as candidate solutions for the next EDA generation.
 \item Both learning and sampling should be efficient. That is, the computational time required to train and sample the model should be tractable both the number of variables, and in the number of training examples.
\end{itemize}    

Previous work has shown that generative neural networks can lead to competitive performance. \cite{Probst2014}  use a Restricted Boltzmann Machine (RBM) in an EDA and show that RBM-EDA can achieve competitive performance to state-of-the art EDAs, especially in terms of computational complexity of the CPU time.  \cite{probst2015dae-eda-better-arxiv} use another type of neural network, a Denoising Autoencoder (DAE) in an EDA. DAE-EDA achieves superior performance when used on problems which can be decomposed into independent subproblems.
In general, neural-network inspired probabilistic models can often be parallelized on massively parallel systems such as graphics processing units (GPUs) \cite{Probst2014a}.

In this paper, we focus on Deep Boltzmann Machines (DBMs) \cite{salakhutdinov2009deep}. DBMs are \textit{deep} models in the sense that they use multiple layers $\mathbf{h}^j$, $j=1\dots d$ of hidden (latent) neurons\footnote{We use the following notation:   $x$ denotes a scalar value, $\mathbf{x}$ denotes a vector of scalars, $\mathbf{X}$ denotes a matrix of scalars}. A DBM models the joint probability distribution $P(\mathbf{v},\mathbf{h}^1,\dots,\mathbf{h}^d)$ of the training data $\mathbf{v}$ and the hidden neurons.

Deep architectures are particularly interesting, because they are able to model problems on multiple layers of abstraction. A deep model is usually composed of multiple layers of computational units (e.g., neurons). The concepts modeled by each layer becomes more abstract with the layer's depth \cite{Bengio2009deep,lecun2015deep}. An intuitive example is a deep neural network that learns to model images of faces \cite{lee2009convolutional}: Neurons on the first hidden layer learn to model individual edges and other shapes. Units on deeper layers compose these edges to form higher-level features, like noses or eyes. Again, by combining theses mid-level representations, neurons in the deepest layers can compose complete faces. Many real-world problems like image classification possess this kind of hierarchic structure with various layers of abstraction. Deep models have recently gained much attention, as they were able to yield superior results for various real-world problem domains \cite{lecun2015deep}.

We implement a DBM and use it within an EDA to solve combinatorial optimization problems. We test DBM-EDA on the simple onemax problem, concatenated deceptive trap functions, NK landscapes and the HIFF function. We compare the results the state-of-the-art multivariate Bayesian Optimization Algorithm (BOA, see \cite{Pelikan1999,Pelikan2005}), and publish the source code of all experiments\footnote{See https://github.com/wohnjayne/eda-suite/ for the complete source code}

Section \ref{dbm}, introduces DBMs. Section \ref{experiments} describes benchmark problems, experimental setup, and presents the results. We discuss the results and conclude the paper in section \ref{conclusion}.

\begin{algorithm}[bp]
\caption{Estimation of Distribution Algorithm}
\label{alg-eda}
\begin{algorithmic}[1]
\STATE \textbf{Initialize} Population $P$
\WHILE {not converged}
\STATE    $P_{parents}$ $\leftarrow$ \textbf{Select} high-quality solutions from $P$ based on their fitness
\STATE    $M$ $\leftarrow$ \textbf {Build} a model  estimating the  (joint) probability distribution of $P_{parents}$ 
\STATE    $P_{candidates}$ $\leftarrow$ \textbf{Sample} new candidate solutions from $M$
\STATE    $P$ $\leftarrow$ $P_ {parents}\cup P_ {candidates}$
\ENDWHILE
\end{algorithmic}
\end{algorithm}

\section{Deep Boltzmann Machines}
\label{dbm}
\begin{figure}
\begin{center}
\centerline{\includegraphics[width=0.4\columnwidth]{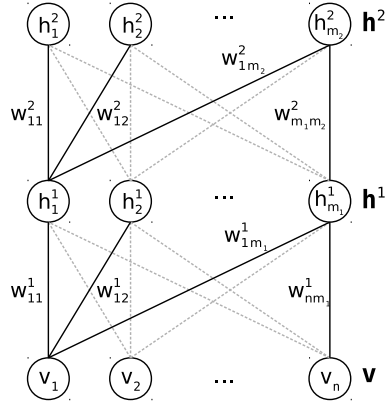}}
\caption{A Deep Boltzmann Machine with two hidden layers $\mathbf{h}^1,\mathbf{h}^2$ as a graph. The visible neurons $v_i$ ($i\in{1..n}$) can hold a data vector of length $n$ from the training data. In the EDA context, $\mathbf{v}$ represents decision variables. The hidden neurons $h^1_j$ ($j \in{1\dots m_1}$) and $h^2_k$ ($k \in{1\dots m_2}$) represent $m_1$ first-level and $m_2$ second-level features, respectively.}
\label{fig-dbm}
\end{center}
\end{figure}
DBMs are special types of Boltzmann Machines, one of the fundamental types of neural networks \cite{ackley1985learning,salakhutdinov2009deep}. A DBM has a visible layer $\mathbf{v} \in [0,1]^n$ and $d$ hidden layers $\mathbf{h}^j \in [0,1]^{m_j}$. Each layer consists of binary units, and learns a non-linear representation of the data on the layer below. Hence, upper layers will learn more abstract concepts.

Here, we focus on DBMs with two hidden layers $\mathbf{h}^1$ and $\mathbf{h}^2$ (see Figure \ref{fig-dbm}). 
The layers of neurons are connected by symmetric weights. The weight matrix $\mathbf{W}^1$ of size ${n*m_1}$ connects the visible layer to the first hidden layer.
Weight $w^{1}_{ij}$ therefore connects $v_i$ to $h^1_j$. 
Accordingly, weight matrix $\mathbf{W}^{2}$ of size $m_1*m_2$ connects hidden layer $\mathbf{h}^1$ to hidden layer $\mathbf{h}^2$. There are no connections within the layers.

From Boltzmann Machines, DBMs inherit the concept of a scalar \textit{energy} associated with each configuration of its neurons. The energy of the state $\{\mathbf{v},\mathbf{h}^1,\mathbf{h}^2\}$ is defined as
\begin{equation}
\label{eq:energy}
E(\mathbf{v},\mathbf{h}^1,\mathbf{h}^2;\theta)=-\mathbf{v}^\top \mathbf{W}^{1}\mathbf{h}^1 - \mathbf{h}^{1\top} \mathbf{W}^2 \mathbf{h}^2
\end{equation}
where $\theta=\{ \mathbf{W}^1, \mathbf{W}^2\}$ are the model's parameters\footnote{We omit the bias terms for brevity, see \cite{salakhutdinov2012better}.}.
The probability of a particular configuration of the visible neurons $\mathbf{v}$ under the model is
\begin{equation}
\label{eq:probability}
P(\mathbf{v};\theta)=\frac{1}{Z(\theta)}\sum_{\mathbf{h}^1,\mathbf{h}^2}{exp(-E(\mathbf{v},\mathbf{h}^1,\mathbf{h}^2;\theta))}.
\end{equation}
$Z(\theta)$ is the partition function which normalizes the probability by summing over all possible configurations.  

The \textit{conditional} probabilities of the neurons, given the activations of their neighboring layers, is easier to calculate.
Each neuron $v_i$ calculates its conditional probability to be active as
\begin{equation}
\label{eq:pv}
P(v_i=1|\mathbf{h}^1)=\text{sigm}(\sum_{j}{h^1_j*w^{vh_1}_{ij}}),
\end{equation}
with $\text{sigm}(x)$ being the logistic function $\text{sigm}(x)=\frac{1}{1+e^{-x}}$. The neurons in hidden layer $h^1$ calculate their probability to be active as

\begin{equation}
\label{eq:ph1}
P(h^1_j=1|\mathbf{v},\mathbf{h}^2)=\text{sigm}(\sum_{i}{v_j*w^1_{ij}}+\sum_{k}{h^2_k*w^2_{jk}}),
\end{equation}
and the neurons in hidden layer $h^2$ calculate  their probability to be active as
\begin{equation}
\label{eq:ph2}
P(h^2_k=1|\mathbf{h}^1)=\text{sigm}(\sum_{j}{h^1_j*w^{2}_{jk}}).
\end{equation}
\subsection{Training a DBM}
\begin{algorithm}
\caption{Pseudo code for pre-training a DBM with RBMs}
\label{alg-dbm-pre}
\begin{algorithmic}[1]
\STATE \textbf{Initialize} $\mathbf{W}^1,\mathbf{W}^2$ to small, random values
\STATE  $\mathbf{W}^1\leftarrow$ Train $\text{RBM}_1$ modeling $P(\mathbf{v},\mathbf{h}^1;\mathbf{W}^1)$,
\STATE $\quad\text{using } P_\text{Parents} \text{ as training set}$
\STATE  $\mathbf{W}^2\leftarrow$ Train $\text{RBM}_2$ modeling $P(\mathbf{h}^1,\mathbf{h}^2;\mathbf{W}^2)$,
\STATE $\quad\text{using samples from } P(\mathbf{H}^1|P_\text{Parents};\mathbf{W}^1) \text{ as defined by RBM}_1\text{ as training set}$
\end{algorithmic}
\end{algorithm}
\begin{algorithm}
\caption{Pseudo code for fine-tuning a DBM}
\label{alg-dbm-fine}
\begin{algorithmic}[1]
\STATE \textbf{Set} $0<\alpha<1$, e.g. $\alpha=0.1$
\STATE \textbf{Initialize} DBM's parameters with pre-trained $\mathbf{W}^1$ and $\mathbf{W}^2$
\STATE \textbf{Initialize} fantasy particles $\hat{\mathbf{v}}$ $\hat{\mathbf{h}}^1$ and $\hat{\mathbf{h}}^2$ randomly
\WHILE {not converged}
    \FOR{\textbf{each} example in the training set}
        \STATE \textit{---  Positive phase ---}
        \STATE $\mathbf{v} \leftarrow$ set $\mathbf{v}$ to the current training example
        \STATE $\mathbf{h}^1 \leftarrow$ set to $P(\mathbf{h}^1|\mathbf{v})$,
        \STATE \quad ignoring input from $\mathbf{h}^2$
        \STATE $\mathbf{h}^2 \leftarrow$ set to $P(\mathbf{h}^2|\mathbf{h}^1)$
        \STATE \textit{run mean field approximation:}
        \FOR{a small number of iterations (e.g. ten)}
            \STATE $\mathbf{h}^1 \leftarrow$ set to $P(\mathbf{h}^1|\mathbf{v},\mathbf{h}^2)$
            \STATE $\mathbf{h}^2 \leftarrow$ set to $P(\mathbf{h}^2|\mathbf{h}^1)$
        \ENDFOR
        \STATE \textit{--- Negative phase ---}
        \FOR{a small number of iterations (e.g. five)}
            \STATE $\hat{\mathbf{h}}^1 \leftarrow$ activate stochastically with $P(\hat{\mathbf{h}}^1|\hat{\mathbf{v}},\hat{\mathbf{h}}^2)$
            \STATE $\hat{\mathbf{h}}^2 \leftarrow$ activate stochastically with $P(\hat{\mathbf{h}}^2|\hat{\mathbf{h}}^1)$
            \STATE $\hat{\mathbf{v}} \leftarrow$ activate stochastically with $P(\hat{\mathbf{v}}|\hat{\mathbf{h}}^1)$
        \ENDFOR
        \STATE \textit{--- Calculate and apply gradient ---}
        \STATE $\delta_\text{pos}=\frac{\partial E(\mathbf{v},\mathbf{h}^1,\mathbf{h}^2;\theta)}{\partial\theta}$
        \STATE $\delta_\text{neg}=-\frac{\partial E(\hat{\mathbf{v}},\hat{\mathbf{h}}^1,\hat{\mathbf{h}}^2;\theta)}{\partial\theta}$
        \STATE $\theta:=\theta-\alpha*(\delta_\text{pos}+\delta_\text{neg})$
    \ENDFOR
\ENDWHILE
\end{algorithmic}
\end{algorithm}
In the training phase, the parameters $\theta$ of the DBM have to be adjusted such that the model approximates the probability distribution of the training data. This could, in principle, be done by using the general training procedure of Boltzmann Machines \cite{ackley1985learning}. However, this would be computationally intractable. Hence, a greedy, layer-wise pretraining is used to initialize the parameters to some sensible values, and subsequently perform parameter fine-tuning on the complete DBM.

During the pretraining phase, we consider the DBM to be a stack of Restricted Boltzmann Machines (RBMs) (see Algorithm \ref{alg-dbm-pre}). RBMs are similar to DBMs, but only have a single layer of hidden neurons. 
Contrastive divergence is a tractable learning algorithm to train RBMs \cite{Hinton2002}.
Specifically, we consider $\mathbf{v}$ and $\mathbf{h}^1$ to be an RBM, with $\mathbf{W}^1$ as its parameters. We then train the resulting RBM to model the probability distribution of the training data, using contrastive divergence (see e.g. \cite{Hinton2002,Hinton2006} for details). Subsequently, we train a second RBM, consisting of $\mathbf{h}^1,\mathbf{h}^2 \text{ and } \mathbf{W}^2$. The training data for the second RBM consists of the activations of the first RBM's hidden layer $\mathbf{h}^1$ on the original training data\footnote{There is a small adjustment in the training procedure for the RBMs. In practice, the weight matrices $\mathbf{W}^1$ and $\mathbf{W}^2$ are multiplied by constant factors, see \cite{salakhutdinov2012better} for details. }.

Once both RBMs have been trained, we use their parameters $\mathbf{W}^1$ and $\mathbf{W}^2$ as an initialization for the parameters $\theta$ of a single DBM.
Note that the pretraining does \textit{not} adjust the weights s.t. the DBM's conditional probability distribution $P(\mathbf{v}|\mathbf{h}^1,\mathbf{h}^2;\theta)$ approximates the probability distribution of the training data. 

In order to achieve this, we fine-tune the DBM (see Algorithm \ref{alg-dbm-fine}). This can be done using gradient descent algorithm to modify $\theta$. 
The general idea of the algorithm is as follows: In order to increase the probability of a training data point $\mathbf{v}$ under the model, the energy $E(\mathbf{v},\mathbf{h}^1,\mathbf{h}^2;\theta)$ of this configuration has to be decreased (see Equation \ref{eq:probability}). At the same time, the probability of all other configurations has to be lowered, by increasing their energies contained in the partition function $Z$. 

The gradient hence contains two terms. The first term (positive gradient) increases the probability of the current sample under the model, the second term (negative gradient) decreases the probabilities of all other configurations in Z.

The negative gradient is calculated by running a separate Markov chain of \textit{fantasy particles} in the model.
Specifically, all neurons are first initialized randomly. Then, the following two steps are repeated: First, $\mathbf{h}^1$ is activated stochastically, with probabilities $P(\mathbf{h}^1|\mathbf{v},\mathbf{h}^2)$ as in Equation \ref{eq:ph1}. Then, $\mathbf{v}$ and $\mathbf{h}^2$  are activated stochastically, with $P(\mathbf{v}|\mathbf{h}^1)$ as in Equation \ref{eq:pv}, and $P(\mathbf{h}^2|\mathbf{h}^1)$ as in  Equation \ref{eq:ph2}. If the parameter updates are small enough, and the Markov chain is allowed to run a couple of steps between each update (e.g. five steps), the samples will come from the chain's equilibrium distribution. The current state of the fantasy particle is then used to calculate the negative gradient.

The positive term could be approximated by running the same Markov chain as above, but with $\mathbf{v}$ clamped to the current training example. However, this would result in running the Markov chain for many steps, for each training example. Instead, the positive gradient can be approximated using a mean-field approach. First, neurons in $\mathbf{v}$ are set according to the current training example. Then, $P(\mathbf{h}^1)$ and, subsequently, $P(\mathbf{h}^2|P(\mathbf{h}^1))$ are calculated, as in Equations \ref{eq:ph1} and \ref{eq:ph2}. Then, for a small number of steps (e.g. ten steps), $P(\mathbf{h}^1|\mathbf{v},P(\mathbf{h}^2))$ and $P(\mathbf{h}^2|P(\mathbf{h}^1))$ are calculated repeatedly. The resulting configuration of $\mathbf{v}$, $P(\mathbf{h}^1)$ and $P(\mathbf{h}^2)$ is treated as a positive example, and used for the calculation of the positive gradient.
The DBM's parameters $\theta$ are then updated in the direction of the total gradient.

For a more detailed description of the training algorithm, see \cite{salakhutdinov2012better}.
\subsection{Sampling a DBM}
The DBM can be sampled by initializing all neurons to random values, and running the same sampling chain used to retrieve fantasy particles for the negative gradient. In the case of DBM-EDA, we initialize $v$ with the $P_\text{Parents}$, and run the chain for 25 iterations.
\section{Experiments}
We evaluate DBM-EDA using a set of standard benchmark problems. We compare the results of DBM-EDA to those of the state-of-the-art multivariate BOA.
\label{experiments}
\subsection{Test Problems}
We evaluate DBM-EDA on onemax, concatenated deceptive traps, NK landscapes and the HIFF function. All four are standard benchmark problems. Their difficulty depends on the problem size, i.e., problems with more decision variables are more difficult. Furthermore, the difficulty of concatenated deceptive trap functions and NK landscapes is tunable by a parameter. Apart from the simple onemax problem, all problems are composed of subproblems, which are either deceptive (traps), overlapping (NK landscapes), or hierarchical (HIFF), and therefore multimodal. 

The onemax problem assigns a binary solution $\mathbf{x}$ of length $l$ a fitness value $f=\sum^l_{i=1} x_i$, i.e., the fitness of $\mathbf{x}$ is equal to the number of ones in $\mathbf{x}$. The onemax function is rather simple. It is unimodal and can be solved by a deterministic hill climber. 

Concatenated deceptive traps are tunably hard, yet separable test problems \cite{deb1993analyzing}. Here, a solution vector $\mathbf{x}$ is divided into $l$ subsets of size $k$, with each one being a deceptive trap. Within a trap, all bits are dependent on each other but independent of all other bits in $\mathbf{x}$. Thus, the fitness contribution of the traps can be evaluated separately and the total fitness of the solution vector is the sum of these terms.
In particular, the assignment $\mathbf{a}=\mathbf{x}_{i:i+k-1}$  (i.e., the $k$ bits from $x_i$ to $x_{i+k-1})$\footnote{The $k$ variables assigned to trap $l$ do not have to be adjacent, but can be at any position in $x$.} leads to a fitness contribution $F_l$ as
$$
F_l(\mathbf{a}) = \begin{cases} k &\mbox{if } \sum_{i}{a_i}=k, \\
k-(\sum_{i}{a_i}+1) & \mbox{otherwise.}
\end{cases}
$$
In other words, the fitness of a single trap increases with the number of zeros, except for the optimum of all ones.

NK landscapes are defined by two parameters $n$ and $k$ and $n$ fitness components $f_{i}, i\in\{1\,\dots,n\}$ \cite{kauffman1989nk}. A solution vector $\mathbf{x}$ consists of $n$ bits. The bits are assigned to $n$ overlapping subsets, each of size $k+1$. The fitness of a solution is the sum of $n$ fitness components.
Each component $f_{i}$ depends on the value of the corresponding variable $x_i$ as well as $k$ other variables. Each $f_{i}$ maps each possible configurations of its $k+1$ variables to a fitness value. The overall fitness function is
$$
f(\mathbf{x})=1/n\sum_{i=1}^nf_i(x_i,x_{i1},\ldots,x_{iK}).
$$ 
\noindent Each decision variable usually influences several $f_i$. These dependencies between subsets make NK landscapes non-separable, i.e., in general, we cannot solve the subproblems independently. The problem difficulty increases with $k$. $k=0$ is a special case where all decision variables are independent and the problem reduces to a unimodal onemax. We use instances of NK landscapes with known optima from \cite{Pelikan2008techreport2}.

The  Hierarchical If-and-only-if (HIFF) function \cite{watson1998modeling} is defined for solutions vectors of length $n=2^l$ where $l\in\mathbb{N}$ is the number of layers of the hierarchy. It uses a mapping function $M$ and a contribution function $C$, both of which take two inputs.
The mapping function takes each of the $n/2$ blocks of two neighboring variables of level $l=1$, and maps them onto a single symbol each. An assignment of $00$ is mapped to $0$, $11$ is mapped to $1$ and everything else is mapped to the null symbol '-'. The concatenation of $M$'s outputs on level $l$ is used as M's input for the next level $l+1$ of the hierarchy, i.e., if level $l=1$ has $n$ variables, level $l=2$ has $n/2$ variables.
On each level, $C$ assigns a fitness to each block of two variables. The assignments $00$ and $11$ are both mapped to $2^l$, everything else to $0$. The total fitness is the sum of all blocks' contributions on all levels.
In other words, a block contributes to the fitness on the current level if both variables in a block have the same assignment. However, only if neighboring blocks agree on the assignment, they will contribute to the fitness on the next level, which is why HIFF is a difficult problem. HIFF has two global optima, the string of all ones, and the string of all zeros. 

\subsection{Experimental Setup}
\label{setup}
We use several instances of the test problems. For each instance and algorithm, we test multiple population sizes between 100 and 16,000\footnote{$\text{popsize}\in\{100;200;300;400;500;1,000;1,500; \text{ and } 2,000 \text{ to } 16,000 \text{ (increment 1000)}\}$}.

We run 20 instances for each population size.
In each run, the EDAs are allowed to run for up to 150 generations. We terminate a run if there is no improvement in the best solution for more than 50 generations. 
These settings make it very unlikely that a run is terminated prematurely, i.e., before convergence.
Both DBM-EDA and BOA use tournament selection without replacement of size two \cite{Miller95geneticalgorithms}.
Note that all test problems, with the exception of NK landscapes, have the string of all ones as their global optimum, for any problem size. To avoid any possible model-induced bias towards solutions with ones or zeros, we generate a random matrix $R\in[0,1]^{n*m}$ of ones and zeros for each run. In each generation, we apply the following operations. Before training, we set $\text{trainingData}=P_\text{parents}\oplus R$, with $\oplus$ being a logical XOR. After sampling we set $P_\text{candidates}=\text{modelSamples}\oplus R$.
and  after sampling. These operations are transparent to correlations between variables.

We use standard values for all hyper-parameters governing the DBM's learning and sampling procedures. All hyper-parameters, and further details of the learning process such as momentum or weight decay are available in a configuration file along with the source code (see git repository on github.com). 

The algorithms are implemented in Matlab/Octave and executed using Octave V3.2.4 on a on a single core of an AMD Opteron 6272 processor with 2,100 MHz.
For the DBM, we used the source code provided by \cite{salakhutdinov2009deep}.

\subsection{Results}
\label{results}
Table \ref{table:results} shows results for DBM-EDA and BOA on the Onemax problem (50, 75, 100, and 150 bit problem size), concatenated 4-Traps (40, 60, and 80 bit), concatenated 5-Traps (25, 50, and 100 bit), NK landscapes with $k \in \{4,5\}$ (30 and 34 bit, two instances each) and the HIFF function (64 and 128 bit).

We report the population sizes, the average number of unique fitness evaluations, and the average CPU times that each algorithm needed to solve the respective problem instance to optimality in at least 50\% and 90\% of the runs.

First, we concentrate on the number of fitness evaluations, and on the results for solving at least 50\% of the runs (left three result columns of table \ref{table:results}). For the simple onemax problem, DBM-EDA needs slightly less fitness evaluations than BOA. For the concatenated deceptive traps, the results are mixed. BOA  finds the optimal solutions to the 4-Trap problems faster than DBM-EDA, while DBM-EDA seems to be more competitive for the harder 5-Traps problem.
For the NK landscapes, DBM-EDA needs less fitness evaluations than BOA in six out of eight instances. On the HIFF problem, DBM-EDA's performance is clearly inferior to BOA: it needs much more fitness evaluations on the 64 bit instance, and is unable to find the optimal solution in at least 50\% of runs for the 128 bit instance. This is surprising, given that HIFF is a hierarchical problem, and the DBM is a hierarchical model. In theory, the DBM should have been able to find the building blocks for HIFF on the lower layer of its representation, and recombine them on the higher layers. 

We now look at the results for solving at least 90\% of runs (right three result columns of table \ref{table:results}). With the exception of the small onemax instances, and the concatenated 5-Traps problem, DBM-EDAs performance is inferior to BOA. In addition to the larger HIFF instance, DBM-EDA is unable to solve four of the NK landscape instances.
In other words, while DBM-EDA was relatively competitive when the goal was to solve at least 50\% of the instances to optimality, it seems to be less able to provide \textit{reliable} convergence to the global optimum. A drastic example is the 150 bit instance of the simple onemax problem: DBM-EDA needs a population size of 200 to find the optimum in at least 50\% of runs, but 6000 to find it in at least 90\% of the runs. This behavior has also been observed in another EDA based on neural networks (DAE-EDA, see \cite{probst2015dae-eda-arxiv}\footnote{Note that this behavior is not so pronounced in later versions of DAE-EDA, which use the same model, but a different parametrization of the learning phase (publication \cite{probst2015dae-eda-better-arxiv} in preparation).}). 

Second, we look at the CPU times required to solve the problem. For most instances, DBM-EDA is faster than BOA. Note that the direct comparison of CPU times is not entirely fair for BOA. In a more efficient programming language instead of a script-based language like Matlab/Octave, BOA's speedup is significantly higher than the one of DAE-EDA. However, neither is Matlab/Octave the best programming language for DBM-EDA: Almost every recent implementation of neural networks is parallelized on graphics processing units (GPU), which, in turn, speeds up training and sampling these models considerably (see e.g. \cite{srivastava2014dropout,krizhevsky2012imagenet,Sutskever2014Sequence}). Parallelizing multivariate EDAs such as BOA is well possible, however the speedups are often single- or double-digit, even on GPUs (see e.g. \cite{ovcenavsek2000parallel,munawar2009theoretical}).
In contrast, parallelizing EDAs using neural networks can make proper use of modern GPU hardware and yield very high speedups: \cite{Probst2014a} report speedups of up to 200$\times$, against optimized CPU code, for RBM-EDA, which uses a neural network model that is closely related to the DAE.
Hence, it is reasonable to assume that an efficient GPU-based implementation of DBM-EDA will still be fast.

However, other neural network based EDAs such as DAE-EDA are computationally less expensive. Hence, they need considerably less time for solving the same benchmark instances to optimality (\cite{probst2015dae-eda-arxiv,probst2015dae-eda-gecco,probst2015dae-eda-better-arxiv}).

\begin{table}
\scriptsize
\begin{tabular}{| c | @{\hskip 0.1cm}c@{\hskip 0.1cm} ||c@{\hskip 0.1cm}|@{\hskip 0.05cm}r@{\hskip 0cm} l@{\hskip 0.1cm}|@{\hskip 0.05cm}r@{\hskip 0cm} l@{\hskip 0.1cm}|c@{\hskip 0.1cm}|@{\hskip 0.05cm}r@{\hskip 0cm} l@{\hskip 0.1cm}|@{\hskip 0.05cm}r@{\hskip 0cm} l@{\hskip 0.1cm}|}
\hline
\multirow{2}{1.6cm}{\textbf{Problem}}&\multirow{2}{0.9cm}{\textbf{Algo}}&\multicolumn{10}{c|}{\shortstack{\textbf{Average results}\\Population size such that optimum is found}}
\\

&&\multicolumn{5}{c|}{ in $\geq$50\% of runs}
&\multicolumn{5}{c|}{ in $\geq$90\% of runs}
\\

&&PopSize&\multicolumn{2}{c|}{Unique Evals}&\multicolumn{2}{c|}{Time (sec)}&PopSize&\multicolumn{2}{c|}{Unique Evals}&\multicolumn{2}{c|}{Time (sec)}\\
\hline 
\multirow{2}{1.6cm}{ONEMAX50}&BOA&125&2,119&$\pm$125&685&$\pm$101&125&2,119&$\pm$125&685&$\pm$101\\
&DBM&100&\textbf{1,700}&$\pm$98&\textbf{442}&$\pm$81&100&\textbf{1,700}&$\pm$98&\textbf{442}&$\pm$81\\
\hline 
\multirow{2}{1.6cm}{ONEMAX75}&BOA&125&2,787&$\pm$158&2,182&$\pm$321&125&2,787&$\pm$158&2,182&$\pm$321\\
&DBM&100&\textbf{2,328}&$\pm$86&\textbf{565}&$\pm$88&100&\textbf{2,328}&$\pm$86&\textbf{565}&$\pm$88\\
\hline 
\multirow{2}{1.6cm}{ONEMAX100}&BOA&250&6,259&$\pm$153&8,967&$\pm$1,016&250&6,259&$\pm$153&8,967&$\pm$1,016\\
&DBM&200&\textbf{5,592}&$\pm$195&\textbf{1,641}&$\pm$306&200&\textbf{5,592}&$\pm$195&\textbf{1,641}&$\pm$306\\
\hline 
\multirow{2}{1.6cm}{ONEMAX150}&BOA&250&7,698&$\pm$270&26,867&$\pm$3,380&250&\textbf{7,698}&$\pm$270&\textbf{26,867}&$\pm$3,380\\
&DBM&200&\textbf{6,095}&$\pm$503&\textbf{3,555}&$\pm$283&6,000&212,464&$\pm$6,846&73,024&$\pm$8,317\\
\hline 
\multirow{2}{1.6cm}{4-Traps 40~bit}&BOA&500&\textbf{8,682}&$\pm$429&1,894&$\pm$323&1,000&\textbf{13,673}&$\pm$758&2,728&$\pm$297\\
&DBM&2,000&34,561&$\pm$6,168&2,034&$\pm$880&3,000&47,231&$\pm$5,712&\textbf{2,201}&$\pm$312\\
\hline 
\multirow{2}{1.6cm}{4-Traps 60~bit}&BOA&500&\textbf{12,152}&$\pm$518&6,797&$\pm$927&1,000&\textbf{20,236}&$\pm$1,362&10,604&$\pm$1,707\\
&DBM&4,000&89,495&$\pm$4,723&\textbf{5,481}&$\pm$416&5,000&104,967&$\pm$12,482&\textbf{6,793}&$\pm$686\\
\hline 
\multirow{2}{1.6cm}{4-Traps 80~bit}&BOA&1,000&\textbf{26,377}&$\pm$780&26,871&$\pm$3,906&2,000&\textbf{43,777}&$\pm$1,695&43,935&$\pm$4,994\\
&DBM&6,000&153,278&$\pm$6,149&\textbf{13,271}&$\pm$1,295&6,000&153,278&$\pm$6,149&\textbf{13,271}&$\pm$1,295\\
\hline 
\multirow{2}{1.6cm}{5-Traps 25~bit}&BOA&1,000&11,032&$\pm$877&1,023&$\pm$245&1,500&14,924&$\pm$1,028&1,384&$\pm$211\\
&DBM&1,000&10,368&$\pm$2,125&\textbf{444}&$\pm$108&1,500&13,291&$\pm$2,471&\textbf{566}&$\pm$108\\
\hline 
\multirow{2}{1.6cm}{5-Traps 50~bit}&BOA&3,000&47,904&$\pm$3,120&20,199&$\pm$2,704&3,000&47,904&$\pm$3,120&20,199&$\pm$2,704\\
&DBM&3,000&51,367&$\pm$19,948&\textbf{3,168}&$\pm$2,087&4,000&49,886&$\pm$11,933&\textbf{3,060}&$\pm$617\\
\hline 
\multirow{2}{1.6cm}{5-Traps 75~bit}&BOA&4,000&\textbf{90,802}&$\pm$2,712&86,908&$\pm$8,345&6,000&119,044&$\pm$4,353&119,275&$\pm$15,826\\
&DBM&5,000&99,990&$\pm$6,169&\textbf{8,538}&$\pm$1,131&6,000&\textbf{101,107}&$\pm$19,431&\textbf{9,183}&$\pm$1,407\\
\hline 
\multirow{2}{1.6cm}{5-Traps 100~bit}&BOA&6,000&\textbf{151,231}&$\pm$3,207&284,456&$\pm$27,058&8,000&190,011&$\pm$4,664&355,140&$\pm$25,659\\
&DBM&8,000&169,700&$\pm$12,290&\textbf{26,802}&$\pm$6,659&8,000&\textbf{169,700}&$\pm$12,290&\textbf{26,802}&$\pm$6,659\\
\hline 
\multirow{2}{1.6cm}{NK $n=30$, $k=4$, $i=1$}&BOA&500&9,820&$\pm$874&1,364&$\pm$274&2,000&\textbf{32,015}&$\pm$3,094&4,590&$\pm$1,044\\
&DBM&300&\textbf{5,976}&$\pm$941&\textbf{453}&$\pm$88&5,000&69,081&$\pm$9,269&\textbf{2,120}&$\pm$576\\
\hline 
\multirow{2}{1.6cm}{NK $n=30$, $k=4$, $i=2$}&BOA&2,000&37,883&$\pm$3,120&6,753&$\pm$1,551&4,000&\textbf{67,939}&$\pm$6,649&13,360&$\pm$3,445\\
&DBM&2,000&\textbf{30,124}&$\pm$3,941&\textbf{1,508}&$\pm$310&9,000&126,982&$\pm$13,840&\textbf{3,529}&$\pm$680\\
\hline 
\multirow{2}{1.6cm}{NK $n=34$, $k=4$, $i=1$}&BOA&500&11,685&$\pm$788&1,896&$\pm$382&1,000&\textbf{21,546}&$\pm$1,860&3,603&$\pm$625\\
&DBM&400&\textbf{8,823}&$\pm$1,227&\textbf{567}&$\pm$105&4,000&69,125&$\pm$6,476&\textbf{2,807}&$\pm$631\\
\hline 
\multirow{2}{1.6cm}{NK $n=34$, $k=4$, $i=2$}&BOA&2,000&\textbf{41,260}&$\pm$3,544&9,178&$\pm$1,885&5,000&\textbf{88,321}&$\pm$9,272&\textbf{20,377}&$\pm$5,123\\
&DBM&4,000&65,752&$\pm$6,832&\textbf{2,758}&$\pm$525&-&-&&-&\\
\hline 
\multirow{2}{1.6cm}{NK $n=30$, $k=5$, $i=1$}&BOA&250&5,835&$\pm$867&787&$\pm$221&500&\textbf{11,221}&$\pm$644&1,565&$\pm$260\\
&DBM&200&\textbf{4,284}&$\pm$851&\textbf{611}&$\pm$290&1,000&17,617&$\pm$1,601&\textbf{1,045}&$\pm$178\\
\hline 
\multirow{2}{1.6cm}{NK $n=30$, $k=5$, $i=2$}&BOA&500&\textbf{12,122}&$\pm$1,456&1,584&$\pm$350&2,000&\textbf{41,641}&$\pm$5,157&\textbf{6,831}&$\pm$1,541\\
&DBM&2,000&33,963&$\pm$3,124&1,773&$\pm$247&-&-&&-&\\
\hline 
\multirow{2}{1.6cm}{NK $n=34$, $k=5$, $i=1$}&BOA&6,000&130,026&$\pm$7,306&34,671&$\pm$4,232&16,000&\textbf{307,171}&$\pm$24,227&\textbf{86,846}&$\pm$15,431\\
&DBM&6,000&\textbf{106,392}&$\pm$10,966&\textbf{4,153}&$\pm$793&-&-&&-&\\
\hline 
\multirow{2}{1.6cm}{NK $n=34$, $k=5$, $i=2$}&BOA&13,000&245,051&$\pm$30,834&63,222&$\pm$16,462&16,000&\textbf{300,058}&$\pm$42,914&\textbf{84,266}&$\pm$22,365\\
&DBM&10,000&\textbf{192,125}&$\pm$18,928&\textbf{6,752}&$\pm$1,254&-&-&&-&\\
\hline 
\multirow{2}{1.6cm}{HIFF64}&BOA&500&\textbf{11,991}&$\pm$731&7,480&$\pm$1,059&500&\textbf{11,991}&$\pm$731&7,480&$\pm$1,059\\
&DBM&3,000&64,073&$\pm$4,283&7,697&$\pm$843&3,000&64,073&$\pm$4,283&7,697&$\pm$843\\
\hline 
\multirow{2}{1.6cm}{HIFF128}&BOA&1,000&\textbf{35,477}&$\pm$1,537&\textbf{99,782}&$\pm$10,278&1,500&\textbf{51,008}&$\pm$2,387&\textbf{137,617}&$\pm$12,151\\
&DBM&-&-&&-&&-&-&&-&\\
\hline\end{tabular}
\caption{This table shows average results for fitness evaluations and CPU time for DBM-EDA and BOA for the test problems. For each instance and algorithm, we selected the minimal population size which leads to the optimal solution in at least 10 (left three result columns) or 18 (right three result columns) of 20 runs. Bold results are significantly smaller, according to a Wilcoxon signed-rank tests ($p<0.01$, data is not normally distributed)}
\label{table:results}
\end{table}

\begin{figure}
  \begin{center}
   \includegraphics[width=0.25\linewidth]{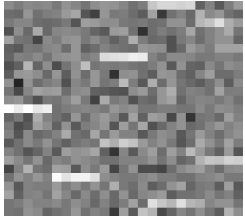}\label{fig:features}
    \caption{\small Visualization of the learned \textbf{weight matrix} $\mathbf{W}^1$ of DBM-EDA optimizing a concatenated 5-trap problem with 30 bits, in a later EDA generation.
    \label{fig:features}
    }
  \end{center}
\end{figure}

\section{Discussion and Conclusion}
\label{conclusion}
We introduced DBM-EDA, an Estimation of Distribution Algorithm which uses a Deep Boltzmann Machine as its probabilistic model. We evaluated the performance of DBM-EDA on multiple instances of standard benchmark problems for combinatorial optimization, and compared the results to the state-of-the-art multivariate Bayesian Optimization Algorithm.

DBM-EDA was able to solve most of the instances. However, its model quality was not competitive to BOA, with the exception of concatenated 5-Trap problems. Specifically, DBM-EDA was often unable to provide reliable convergence to the global optimum, or needed very large population sizes. Correspondingly, a high number of fitness evaluations were required. A reason for the insufficient model quality might be the mean-field approximation of the DBM's training process. Mean-field approximation struggles if the probability distribution being approximated is multimodal. However, this is often the case in the early EDA generations: A sample might resemble configurations of different local optima, perturbed by noise.
Surprisingly, DBM-EDA was unable to solve the larger HIFF instance at all. This is despite the fact that, as a hierarchical model with multiple layers, DBM-EDA should be particularly well-suited for a hierarchical optimization problem like HIFF. 

DBM-EDA' performance the concatenated trap problem with traps of size $k=5$ was superior to BOA. Here, DBM-EDA was able to solve the larger instances (75 and 100 bit) with fewer fitness evaluations. Recall that the problem is particularly difficult, as it is composed of subproblems which are deceptive. We hypothesize that the reason for the good performance is the structure of the DBM: The hidden neurons are conditionally independent (see Equations \ref{eq:ph1} and \ref{eq:ph2}). This matches the fitness function, which is additively decomposable, and separable. Figure \ref{fig:features} shows that neurons in the first hidden layer tend to model global optima to different additive parts of the fitness function. White pixels indicate large positive weight values, black pixels large negative weight values. Each row visualizes the connections between a single hidden neuron (of the first hidden layer) and the 30 problem variables. In the deceptive 5-trap problem, blocks of five adjacent variables have a strong contribution to the fitness, if \textit{all five} variables are equal to one or equal to zero. Each block of five variables is independent of all other blocks.
The figure shows that many hidden neurons strongly influence a single block of problem variables (bright/dark blocks of five adjacent pixels), and are indifferent to most other neurons (mid gray values). The learned representation of the model therefore matches the problem structure.
As the hidden neurons are activated stochastically, samples can comprise local optima of different additive terms of the fitness function, even if this specific combination has not been seen in the training data.  A similar behavior has been observed for DAE-EDA, another EDA using neural networks \cite{probst2015dae-eda-better-arxiv}.

In sum, while it is feasible to use a DBM as an EDA model, the effort for learning the multi-layered DBM model seems not to pay off for the optimization process in a noisy environment. 
There are multiple areas for future research. In the case where only 50\% of the runs were required to find the global optimum, the results for DBM-EDA were quite encouraging. DBM-EDA could be a useful tool, if it could provide reliable convergence. The reasons to why this is currently not the case should be analyzed properly. Also, more work is necessary to understand why DBM-EDA was unable to apply the benefits of its hierarchical model to the HIFF problem.

\bibliographystyle{abbrv}

\end{document}